\documentclass{article}

\usepackage{microtype}
\usepackage{graphicx}
\usepackage{subfigure}
\usepackage{booktabs} 

\usepackage{hyperref}

\usepackage[accepted]{icml2023}

\usepackage{amsmath}
\usepackage{amssymb}
\usepackage{mathtools}
\usepackage{amsthm}
\usepackage{tabularx}
\usepackage{enumitem}
\usepackage{multirow}
\usepackage{multicol}
\usepackage{makecell}

\usepackage[capitalize,noabbrev]{cleveref}

\theoremstyle{plain}

\theoremstyle{definition}

\theoremstyle{remark}

\usepackage[textsize=tiny]{todonotes}

\icmltitlerunning{Test-Time Adaptation with Perturbation Consistency Learning}

\begin{document}

\twocolumn[
\icmltitle{Test-Time Adaptation with Perturbation Consistency Learning}



\icmlsetsymbol{equal}{*}

\begin{icmlauthorlist}
\icmlauthor{Yi Su}{equal,yyy}
\icmlauthor{Yixin Ji}{equal,yyy}
\icmlauthor{Juntao Li}{yyy}
\icmlauthor{Hai Ye}{sch}
\icmlauthor{Min Zhang}{yyy}
\end{icmlauthorlist}

\icmlaffiliation{yyy}{Institute of Computer Science and Technology, Soochow University, China}
\icmlaffiliation{sch}{Department of Computer Science, National University of Singapore, Singapore}

\icmlcorrespondingauthor{Juntao Li}{ljt@suda.edu.cn}

\icmlkeywords{Machine Learning, ICML}

\vskip 0.3in
]



\printAffiliationsAndNotice{\icmlEqualContribution} 

\begin{abstract}
Currently, pre-trained language models (PLMs) do not cope well with the distribution shift problem, resulting in models trained on the training set failing in real test scenarios. To address this problem, the test-time adaptation (TTA) shows great potential, which updates model parameters to suit the test data at the testing time. 
Existing TTA methods rely on well-designed auxiliary tasks or self-training strategies based on pseudo-label.
However, these methods do not achieve good trade-offs regarding performance gains and computational costs.
To obtain some insights about such a dilemma, we take two representative TTA methods, i.e., Tent and OIL, for exploration and find that stable prediction is the key to achieving a good balance.
Accordingly, we propose perturbation consistency learning (PCL), a simple test-time adaptation method to promote the model to make stable predictions for samples with distribution shifts in this paper.
Extensive experiments on adversarial robustness and cross-lingual transferring demonstrate that our method can achieve higher or comparable performance with less inference time over strong PLM backbones and previous state-of-the-art TTA methods.
\end{abstract}

\section{Introduction}
Pre-trained language models (PLMs) have achieved stunning performance on many NLP tasks \cite{devlin2019bert,liu2019roberta,lewis2020bart,raffel2020exploring,brown2020language,radfordimproving,chatgpt}.
Nevertheless, we are still far from achieving a trustworthy NLP model. 
In practical scenarios, the test data inevitably shift in distribution from the training data, such as adversarial attack sample \cite{wang-etal-2022-measure}, cross-lingual \cite{li2021unsupervised}, cross-domain \cite{ramponi2020neural}, and so on \cite{hupkes2022state}.
Unfortunately, most current NLP models are not robust to distribution shifts \cite{ribeiro-etal-2020-beyond,wang-etal-2022-measure}.
Previous work has explored several approaches to improve the performance of PLMs on out-of-distribution data, such as adversarial training \cite{madry2018towards,zhu2019freelb}, data augmentation \cite{zhou-etal-2021-virtual}.
These methods attempt to cover more data distribution during the training phase and reduce the difference in distribution between the training and test data. However, they cannot prevent the occurrence of distribution shifts in the face of dynamically changing real-world scenarios.
Thus, from another perspective, some work \cite{2021Tent,sun2020test,niu2022efficient,yerobust} have attempted to update the model based on test samples to accommodate distribution shifts during the testing phase of the model, called test-time adaptation (TTA).

The core of TTA is how to update the model with unlabelled test data, sacrificing as little inference time as possible for a model that is more robust to distribution shift.
Existing TTA methods can be grouped into two categories: 
(1) based on well-designed auxiliary tasks, such as rotation angle prediction and contrastive learning \cite{sun2020test,liu2021ttt++,bartler2022mt3,gandelsman2022testtime,chen2022contrastive};
(2) based on self-training strategies \cite{2021Tent,niu2022efficient,zhang2022memo,yerobust,goyal2022test}.
The first approach requires training data to train additional auxiliary task modules. However, in practice, we may not have access to training data, and retraining an already deployed model would waste many resources.
The second one circumvents the shortcomings of the first one. Still, the performance of the self-training strategy depends on the quality of the pseudo-labels output by the model, and low-quality pseudo-labels can lead to the accumulation of errors in the iterative self-training process.

To gain more insights, we launch a study on two representative TTA methods, Tent and OIL. We find that Tent uses dropout in the test-time adaptation phase, which outputs soft labels that suffer from instability.
In contrast, OIL uses pseudo-labels derived from the stable inference results of the teacher model.
Although OIL can achieve a more robust model than the other methods, it requires forward propagation several times, resulting in a significant sacrifice of inference time.
Therefore, there are reasons to believe that stable prediction is the key to achieving a good balance regarding performance gains and inference speed.

In this paper, we propose perturbation consistency learning (PCL) to achieve Pareto improvement \cite{liu2022towards} over existing TTA methods.
Expressly, we turn off dropout for the model to have a stable output during the test-time phase, and, in order not to lose the robustness brought by dropout, we add perturbations to the output features to constrain the consistency of the model's predictions for the original and perturbed samples.
To verify the effectiveness of our proposed method, we launch experiments on two distribution shift settings (adversarial attack and cross-lingual) and two tasks (question answering and named entity recognition).
Experimental results demonstrate that our proposed PCL significantly improves the performance of out-of-distribution data upon strong PLMs and achieves higher or comparable performance with less inference time over previous state-of-the-art TTA methods.

Overall, our contributions are shown below:
\begin{itemize}[leftmargin=*]
\setlength{\itemsep}{0pt}
\setlength{\parskip}{0pt}
\item We investigate the reasons for the poor performance of typical fully test-time adaptation in NLP tasks. We find that the unstable output caused by dropout in the test-time stage is the critical factor causing this problem.
\item We propose the perturbation consistency learning (PCL) for TTA in NLP tasks, which turns off dropout in the encoder and constrains the predictive consistency of the model to feature perturbations.
\item Experimental results and analysis on QA and NER tasks under adversarial attack and cross-lingual transfer settings demonstrate that our proposed PCL achieves a better trade-off between performance and efficiency than previous TTA methods.
\end{itemize}

\section{Related Work}
\paragraph{Test-Time Adaptation}
Test-Time Adaptation (TTA) is a new paradigm to deal with the distribution shift between training data and test data.
TTA uses self-supervised signals to update and give prediction results at the inference stage.
It has achieved great success in CV tasks \cite{sun2020test,liu2021ttt++,2021Tent,bartler2022mt3,niu2022efficient,gandelsman2022testtime,zhang2022memo}.
TTA can be divided into test-time training and fully test-time adaptation according to whether or not to modify the target of the training phase and access the training data.
Test-time training usually requires a well-designed self-supervised auxiliary task to adjust the model to the distribution shift.
\citet{sun2020test} utilize the rotation angle prediction as the auxiliary task in the training stage and adapt the model during the test time.
Following this line of thought, a series of works analyze and devise more effective auxiliary tasks
\cite{liu2021ttt++,bartler2022mt3,gandelsman2022testtime,chen2022contrastive}.
On the other line, the fully test time adaptation updates the model with test data using a self-training strategy without modifying the training process.
\citet{2021Tent} propose Tent, which adjusts the model by minimizing entropy.
However, fully test-time adaptation faces performance degradation due to the low quality of pseudo-labels and catastrophic forgetting, so current works \cite{niu2022efficient,zhang2022memo,jang2022test,gong2022note} focus on improving the quality of pseudo-labels and designing more robust tuning methods.
In NLP, TTA has also attracted the attention of many researchers.
\citet{wang2021efficient} learn to combine adapters on low-resource languages at test time to improve sequence labeling tasks.
\citet{banerjee2021self} generate question and answer pairs for training during testing time.
\citet{yerobust} first evaluate the effectiveness of fully test-time adaptation in the question-answering task and introduce a teacher-student paradigm to ensure the quality of pseudo-labels.

\paragraph{Robustness in NLP tasks}
Building robust models is the only way to be trustworthy artificial intelligence, so the field of NLP has been pursuing the robustness of models.
\citet{wang-etal-2022-measure} divide robustness in NLP into two categories, one for adversarial robustness under artificial attacks and one for naturally occurring distribution shifts.
For adversarial robustness, researchers \cite{ebrahimi2018hotflip,alzantot-etal-2018-generating,jia-etal-2019-certified,garg-ramakrishnan-2020-bae,lin2021rockner,ravichander2021noiseqa} have proposed a number of methods for constructing adversarial attack samples to evaluate the model's adversarial robustness.
Adversarial samples can be correctly identified for humans but are confusing for models.
\citet{li2018textbugger} use the Jacobian matrix to decide the word to be modified and modify it by different strategies.
\citet{li2020bert} find vulnerable words with BERT and replace them with top-K words generated by BERT.
In NLP, naturally occurring distribution shifts have rich real-world scenarios.
Thus researchers hope models that are robust to distribution shift to perform well across languages \cite{hu2020xtreme,DBLP:journals/corr/abs-2110-02052,zheng2021consistency,yang2021enhancing,mao2022lessforgetting}, across domains \cite{hendrycks-etal-2020-pretrained,ramponi2020neural,malinin2021shifts}, and on different styles of text.
Many robustness tuning methods are proposed to improve the robustness of NLP models, such as data augmentation \cite{kaushik2019learning,khashabi-etal-2020-bang,chen-etal-2020-mixtext,chen2021hiddencut,zhou-etal-2021-virtual} and adversarial training \cite{miyato2016adversarial,madry2018towards,Zhu2020FreeLB:,wang2021infobert}.

\section{Preliminary}
\subsection{Problem Formulation}
In this paper, we only discuss the fully test-time adaptation setting, in which we cannot access training data.
Suppose we have training data pairs $\{x_s^i, y_s^i\}_{i=1}^{n_s} \in \mathcal{D}_s$ where $x_s^i \in \mathcal{X}_s$ and $y_s^i \in \mathcal{Y}_s$, $x_s^i$ refers to the input and $y_s^i$ refers to the corresponding labels.
The distribution of the training data is $P_s$.
We define an encoder $f(\cdot): \mathcal{X}_s\xrightarrow{} \mathbf{R}^D$ and a classifier $g(\cdot): \mathbf{R}^D\xrightarrow{} \mathbf{R}^C$ where D and C are feature dimensions and the number of classes.
The model is parameterized to $\theta$ after training.
In the test phase, we have test data samples $\{x_t^i\}_{i=1}^{n_t}\in \mathcal{X}_t$. 
Suppose the underlying corresponding labels are $\{y_t^i\}_{i=1}^{n_t}\in \mathcal{Y}_t$. 
The distribution of the test data is $P_{t} \ne P_{s}$.
If we use the trained model $\theta$ to infer the test data, the model's performance will be degraded due to the distribution shifts.
We adapt the model to the test data to avoid performance degradation by optimizing an unsupervised loss during testing $L(x_{t})$.
Some representative fully test-time adaptation methods, such as Tent, EATA use the predicted probability as pseudo labels and minimize the entropy of model predictions:
\begin{equation}
    \label{eq1}
    L(x_{t})= -\sum_c p\left(\hat{y}_{c}|x_t\right) \log p\left(\hat{y}_{c}|x_t\right),
\end{equation}
where $p\left(\hat{y}_c\right)$ is the probability of the c-th category.

\subsection{What limits the effect of TTA in NLP tasks?}
\begin{figure}[t]
    \vskip 0.2in
    \centering
    \includegraphics[scale=0.35]{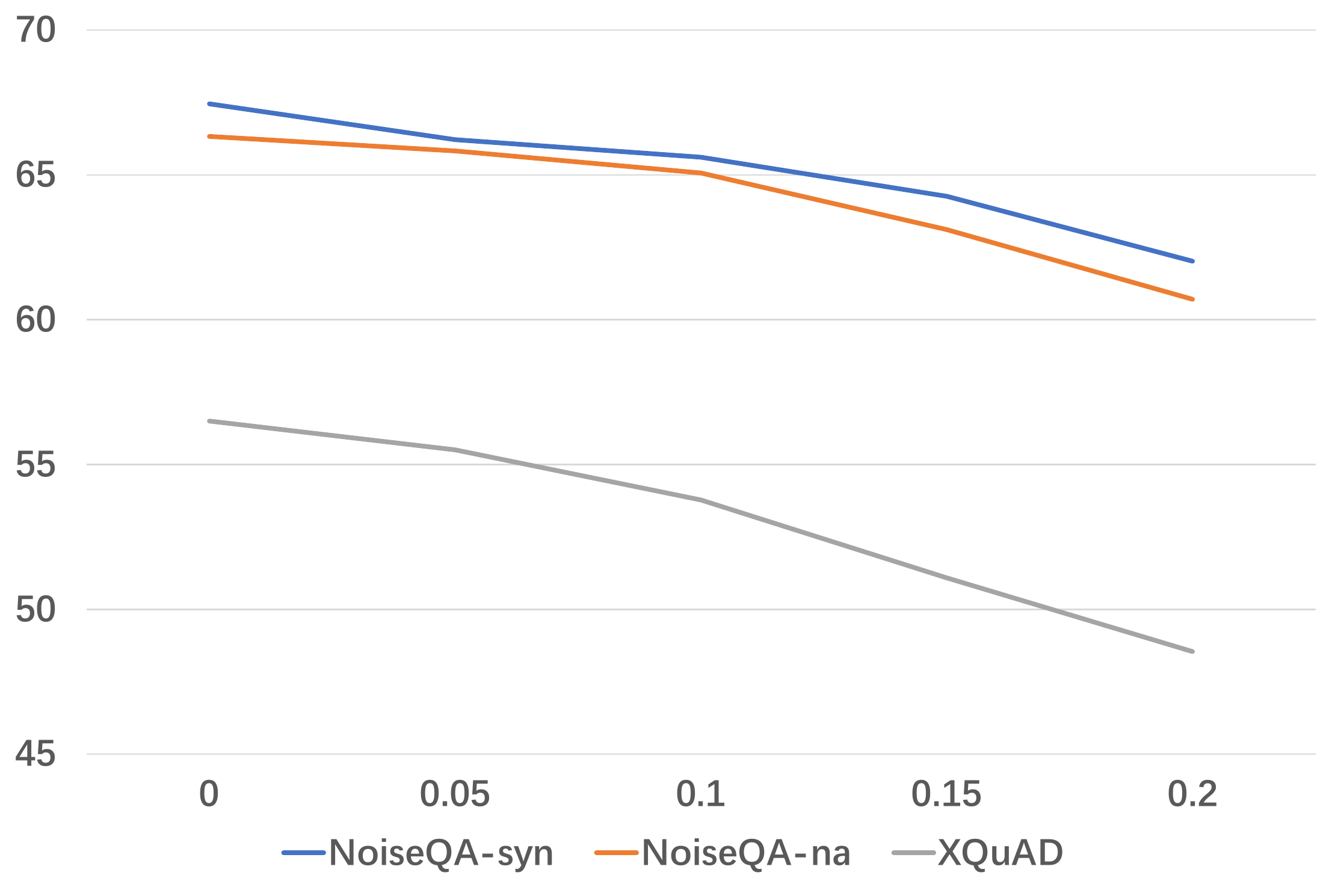}
    \caption{Results (\%) of different dropout rates on NoiseQA-syn, the base model is XLMR-base. Abscissa refers to the dropout rate.}
    \label{fig:closedrop}
    \vskip -0.2in
\end{figure}
In this section, we take Tent \cite{2021Tent} as an example and discuss what limits the effect of TTA.
We choose Tent because it is a typical and strong baseline, and many advanced TTA methods are based on it.
Tent performs very well in CV tasks with ResNet \cite{he2016deep}, but we find in our experiments that Tent does not perform as well when migrating to NLP tasks with pre-trained language models.
In CV tasks, Tent keeps the model in the training mode and updates BatchNorm layers.
The widely used ResNet does not have dropout layers, so the training mode only affects the BatchNorm layer.
However, in NLP tasks, mainstream pre-trained language models all use the Transformer network, including the dropout layer and LayerNorm layer, so the training mode will impact dropout.
In order to increase the robustness of the model during training, Tent uses dropout. However, dropout will also make the model lose some helpful information, resulting in the unstable output of the model.
The unstable output will have two effects on Tent.
Firstly, the output quality will be reduced, thus reducing the model's accuracy.
Secondly, the output will be used as pseudo-labels to guide the update of the model. Low-quality pseudo-labels will make it easier for the model to learn the wrong information.
In contrast, OIL, a TTA method, which achieves performance improvement in NLP tasks, turns off the teacher model's dropout and uses the teacher model's stable prediction to guide the student model.
In order to study the effect of dropout on the performance of Tent, we have done a simple experiment on the NoiseQA-syn dataset \cite{ravichander2021noiseqa}.
The results are shown in Figure \ref{fig:closedrop}.
We use Tent under different dropout rates and find that the performance of Tent will gradually improve as the dropout rate decreases.
Thus, there are reasons to believe that dropout's improvement in model robustness and its instability of output constitutes a contradiction, which limits the performance of fully TTA methods represented by Tent in NLP tasks.

\section{Method}
\begin{figure}[t]
    \vskip 0.2in
    \centering
    \includegraphics[scale=0.5]{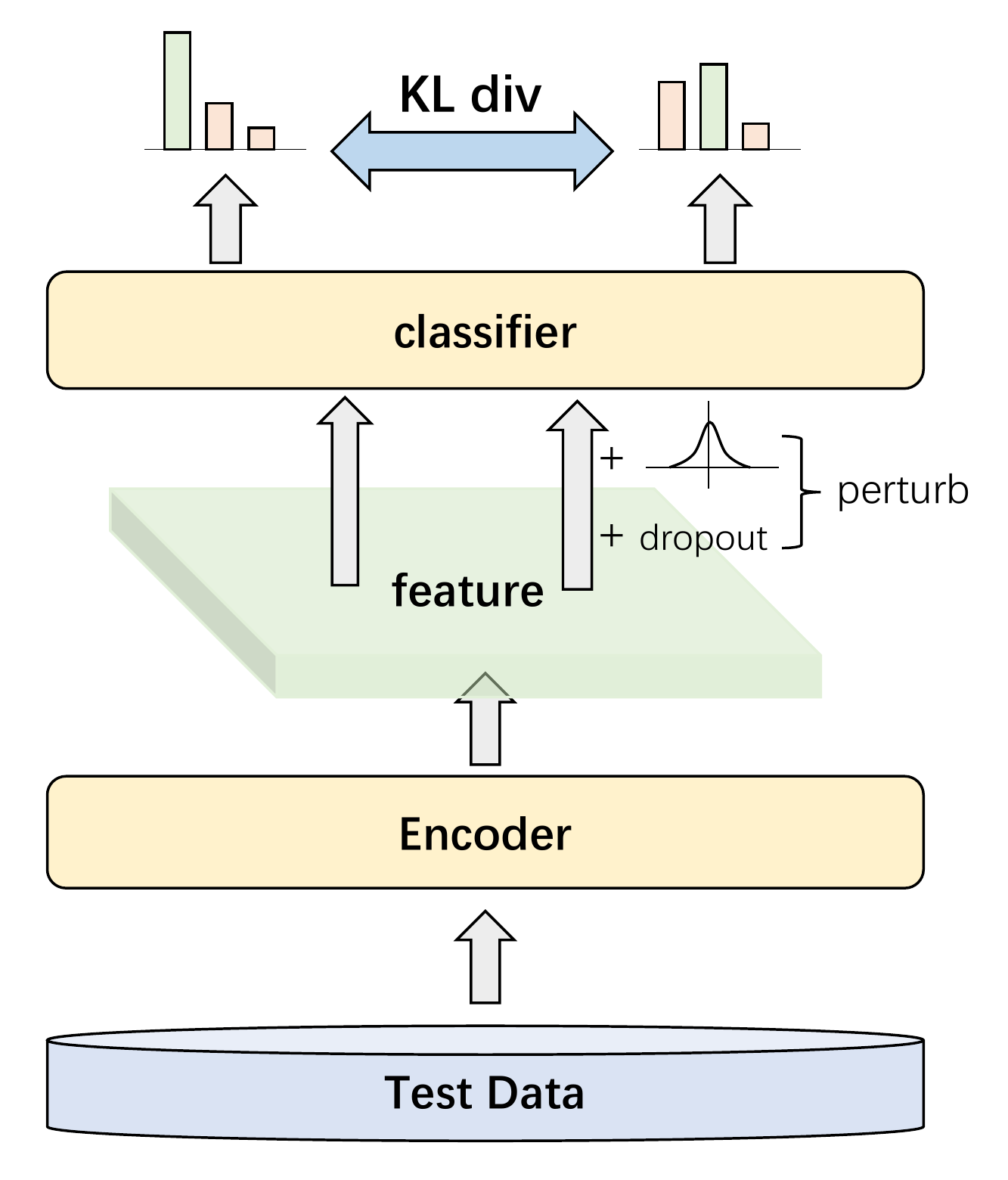}
    \caption{Overview of perturbation consistency learning. The final prediction is made on the original feature. }
    \label{fig:model}
    \vskip -0.2in
\end{figure}
In this section, we introduce our perturbation consistency learning (PCL) method.
Figure \ref{fig:model} shows the overview of our proposed method.
We already have a trained model $f\circ{g}_{\theta}$ and test data $\{x_t^i\}_{i=1}^{n_t}\in \mathcal{X}_t$.
The trained model outputs predicted probability $p(\hat{y}|x_t)=g(f_\theta\left(x_t\right))$.

\subsection{Perturbation Consistency Learning}
As we found in Section 3.2, turning on dropout during the test time phase will lose the features of the sample and cause the quality of the pseudo-label to decrease, which may make the test-time adaptation fail; while turning off dropout may cause over-fitting when optimizing the model \cite{srivastava2014dropout}.
To break this dilemma, we propose perturbation consistency learning by adding slight perturbation in the feature space and encouraging the model to produce consistent prediction between the original and perturbed samples, which is different from dropout consistency regularization \cite{liang2021r}.

Specifically, given the test data $x_{t}$, we feed $x_{t}$ to go through the forward pass of the model and obtain the feature $\boldsymbol{h}$, then we add random perturbation to $\boldsymbol{h}$.
We will introduce our perturbation strategies in Section 4.2.
The classifier will output two predicted probabilities of the original feature $\boldsymbol{h}$ and perturbed feature $\boldsymbol{h}^{\prime}$, denoted as $p(\hat{y}|x_t)$ and $p^{\prime}(\hat{y}|x_t)$.
To make the model robust to feature perturbations, we constrain the predicted probability of the perturbed features to be consistent with the original features.
We use Kullback-Leibler (KL) divergence as the objection to model optimization:
\begin{equation}
\begin{aligned}
\label{eq2}
\mathrm{KL}(p^{\prime} \| p)&=\sum_{c} p^{\prime}(\hat{y}_{c}|x_{t}) \log \frac{p^{\prime}(\hat{y}_{c}|x_{t})}{p(\hat{y}_{c}|x_{t})} \\
&=H(p^{\prime},p)-H(p^{\prime}).
\end{aligned}
\end{equation}
When minimizing the KL divergence, the entropy of $p^{\prime}(\hat{y}|x_{t})$ is maximized, which conflicts with the minimized entropy of Tent, so PCL uses the KL divergence as the objective function $L(x_{t})$ instead of the regularization term of Equation \ref{eq1}.
The overall test-time adaptation algorithm based on our PCL is given in Algorithm \ref{alg:example}.
For mainstream NLP models, the encoder takes up most of the model parameters.
Our PCL method only goes through the encoder once and the classifier twice in forward propagation, so the increase in computation cost compared to Tent is negligible.
\begin{algorithm}[tb]
   \caption{Perturbation Consistency Learning}
   \label{alg:example}
\begin{algorithmic}
   \STATE {\bfseries Input:} Test data $\{x_t^i\}_{i=1}^{n_t}$, the trained model parameter $\theta$.
   \STATE {\bfseries Output:} model parameter $\theta^{\prime}$ and prediction results $\hat{y}$ of test data
   \FOR{batch in test data}
   \WHILE{not converged}
   \STATE 1. feed forward and obtain the output distribution $p(\hat{y}|x_t)$ and $p^{\prime}(\hat{y}|x_t)$;
   \STATE 2. calculate the KL-divergence loss by Equation \ref{eq2};
   \STATE 3. update the model parameters by minimizing Equation \ref{eq2}.
   \ENDWHILE
   \STATE output prediction results $\hat{y}=\mathop{\arg\max}\limits_{c} p(\hat{y}_{c}|x_{t})$
   \ENDFOR
\end{algorithmic}
\end{algorithm}

\subsection{Perturbation Strategies}
We add perturbations to the features to enable the model to remain robust under minor perturbations through the consistency constraint. Overly aggressive perturbations may cause the predicted distributions of the original features and the perturbed features to differ too much, leading to the failure of the consistency constraint, for which we need to choose the perturbation strategy carefully.
In PCL, we use two simple yet effective perturbation strategies: dropout and normal distribution noise.
The feature will be passed to a dropout layer first, then added with the normal distribution noise $\epsilon \sim \mathcal{N}(\mathbf{0},\mathbf{I})$.
This process can be expressed as follows:
\begin{equation}
\boldsymbol{h^{\prime}}=\mathrm{Dropout}(\boldsymbol{h})+\epsilon
\end{equation}
where Dropout means the forward propagation in the dropout layer.
The choice of perturbing methods is diverse.
Theoretically, many other perturb methods can also be effective for PCL, but it should be noted that the perturbation needs to be mild and should not disrupt the original features.

\section{Experiments}
\begin{table*}[htb]
\centering
\caption{\label{tab:qa} Results (\%) on QA for XLMR-base and XLMR-large. We report EM and F1 metrics for each dataset and its average results. We also report the number of samples predicted by the model per second. \textbf{Bold}: the best results. $\mathrm{Star}^{*}$: the second best results. }
\vskip 0.15in
    \begin{tabular}{l l l l l l l l l l l r }
    \toprule
\multirow{2}{*}{\textbf{Models}} & \multicolumn{2}{c}{\textbf{NoiseQA-syn}} & \multicolumn{2}{c}{\textbf{NoiseQA-na}} & \multicolumn{2}{c}{\textbf{XQuAD}} & \multicolumn{2}{c}{\textbf{MLQA}} & \multicolumn{2}{c}{\textbf{Avg.}} & \multirow{2}{*}{\textbf{Speed}}\\
\cmidrule(r){2-3} \cmidrule(r){4-5} \cmidrule(r){6-7} \cmidrule(r){8-9} \cmidrule(r){10-11}
 &\, \textbf{EM} & \, \textbf{F1} & \, \textbf{EM} &\, \textbf{F1} & \, \textbf{EM} & \, \textbf{F1} &\, \textbf{EM} & \, \textbf{F1} & \, \textbf{EM} & \, \textbf{F1} \\
    \midrule
    xlmr-base & 66.64 & 78.67 & 66.25 & 77.91 & 55.59 & 71.42 & 47.14 & 65.27 & 58.91 & 73.32 & 225.78\\
    xTune & 70.95 & 81.52 & 69.75 & 80.66 & 58.78 & 73.75 & 49.87 & 67.76 & 62.34 & 75.92 & 225.78\\
    \midrule
    Tent & 65.61 & 77.06 & 65.06 & 76.38 & 53.78 & 68.96 & 44.18 & 62.30 & 57.16 & 71.18 & 97.21\\
    EATA & 65.23 & 76.73 & 64.65 & 76.05 & 53.95 & 68.85 & 43.98 & 62.02 & 56.95 & 70.91 & 95.92 \\
    OIL & \textbf{68.55} & \textbf{79.65} & \textbf{67.75} & \textbf{78.94} & \textbf{57.63} & \textbf{72.37} & \textbf{47.90} & \textbf{65.54} & \textbf{60.46} & \textbf{74.13} & 12.16  \\
    PCL & $68.44^{*}$ & $79.61^{*}$ & $67.40^{*}$ & $78.69^{*}$ & $56.91^{*}$ & $71.93^{*}$ & $47.38^{*}$ & $65.23^{*}$ & $60.03^{*}$ & $73.87^{*}$ & 104.73 \\
    \midrule
    xTune+Tent & 70.25 & 80.80 & 69.85 & 80.04 & 57.74 & 72.17 & 48.20 & 65.83 & 61.51 & 74.71 & 97.21  \\
    xTune+EATA & 69.83 & 80.42 & 69.48 & 79.80 & 57.54 & 71.90 & 47.97 & 65.55 & 61.21 & 74.42 & 95.92  \\
    xTune+OIL & $71.51^{*}$ & $81.82^{*}$ & $70.35^{*}$ & $80.76^{*}$ & \textbf{59.86} & \textbf{74.38} & \textbf{50.47} & \textbf{68.21} & \textbf{63.05} & \textbf{76.29} & 12.16  \\
    xTune+PCL & \textbf{71.52} & \textbf{81.88} & \textbf{70.68} & \textbf{81.15} & $59.68^{*}$ & $74.06^{*}$ & $50.10^{*}$ & $67.80^{*}$ & $63.00^{*}$ & $76.22^{*}$ & 104.73  \\
    \midrule
    \midrule
    xlmr-large & 65.55 & 79.91 & 64.17 & 78.37 & 63.15 & 78.77 & 53.87 & 72.58 & 61.69 & 77.41 & 77.42 \\
    xTune & 70.11 & 83.17 & 67.20 & 80.54 & 65.00 & 79.91 & 56.30 & 74.33 & 64.65 & 79.48 &77.42 \\
    \midrule
    Tent & 71.08 & 82.97 & 68.57 & 80.65 & 62.27 & 77.77 & 52.19 & 70.94 & 63.53 & 78.08 & 34.32 \\
    EATA & 71.20 & 82.93 & 68.77 & 80.61 & 62.32 & 77.59 & 52.15 & 70.85 & 63.61 & 78.00 & 34.27 \\
    OIL & $71.04^{*}$ & $83.68^{*}$ & $69.59^{*}$ & $81.85^{*}$ & \textbf{64.24} & \textbf{79.40} & \textbf{54.19} & \textbf{72.72} & $64.77^{*}$ & \textbf{79.41} & 4.23  \\
    PCL & \textbf{71.86} & \textbf{83.75} & \textbf{69.73} & \textbf{81.90} & $63.80^{*}$ & $78.97^{*}$ & $54.00^{*}$ & $72.56^{*}$ & \textbf{64.85} & $79.30^{*}$ & 36.69 \\
    \midrule
    xTune+Tent & 75.72 & 86.32 & 73.37 & 83.93 & 65.23 & 79.58 & 55.39 & 73.50 & 67.43 & 80.83 & 34.32  \\
    xTune+EATA & 75.56 & 86.12 & 73.20 & 83.88 & 64.95 & 79.28 & 55.36 & 73.42 & 67.27 & 80.68 & 34.27  \\
    xTune+OIL & $75.94^{*}$ & \textbf{86.72} & $73.52^{*}$ & $84.48^{*}$ & \textbf{66.02} & \textbf{80.34} & $56.21^{*}$ & $74.23^{*}$ & $67.92^{*}$ & \textbf{81.44} & 4.23  \\
    xTune+PCL & \textbf{76.11} & $86.69^{*}$ & \textbf{73.89} & \textbf{84.58} & $65.81^{*}$ & $80.08^{*}$ & \textbf{56.29} & \textbf{74.27} & \textbf{68.03} & $81.41^{*}$ & 36.69  \\
    \bottomrule
    \end{tabular}
\vskip -0.1in
\end{table*}
\subsection{Dataset}
To verify our proposed method's effectiveness, we have conducted experiments on the question answering (QA) and named entity recognition (NER) tasks under adversarial attack and cross-lingual transfer settings.
The datasets we used include the following:

\noindent\textbf{NoiseQA} \cite{ravichander2021noiseqa} defined three kinds of noise based on the SQuAD dataset \cite{rajpurkar2018know}: speech recognition noise, keyboard input noise, and machine translation noise.
Then, the synthetic and natural datasets are constructed for each kind of noise, respectively.

\noindent\textbf{XQuAD} \cite{artetxe2020cross} is a cross-lingual QA dataset. It selects a subset of the SQuAD \cite{rajpurkar2016squad} development set and is translated into ten languages.

\noindent\textbf{MLQA} \cite{lewis2020mlqa} is also a cross-lingual QA dataset. It contains 7 languages but more samples for each language than XQuAD.

\noindent\textbf{RockNER} \cite{lin2021rockner} is a simple and effective method to generate adversarial samples.
The method includes two levels of attack.
At the entity level, it uses other entities of the same semantic type in Wikipedia to replace the target entity.
At the context level, it uses the pre-trained language model to generate word replacement.
The method is applied to the OntoNotes dataset \cite{weischedel2013ontonotes} to generate the adversarial dataset.

\subsection{Baselines}
We leverage the following typical and strong baselines for comparison to calibrate the effectiveness of our PCL:

\noindent\textbf{Tent} \cite{2021Tent} updates the model at test time by entropy minimization. 
The prediction is given immediately, and the updated model will affect the next iteration.

\noindent\textbf{EATA} \cite{niu2022efficient} filters the pseudo-labels of the sample. It only updates the reliable and non-redundant samples.
In addition, EATA restricts the updating of the model so that the model will not forget the knowledge learned in the past.

\noindent\textbf{OIL} \cite{yerobust} uses the teacher-student paradigm to optimize the quality of pseudo-labels, which can prevent the knowledge learned by the model from getting worse.

In addition to the three fully TTA methods mentioned above, we also compare with xTune \cite{zheng2021consistency}, a very strong robustness tuning method, and combine TTA methods with robustness tuning.

\subsection{Setup}
We find that the results of full parameters update are similar to that of only updating LayerNorm parameters.
Thus, to shorten the inference time, all TTA methods only update LayerNorm parameters.
It should be noted that when we reproduce EATA, we do not consider its redundant filtering because different tasks make it difficult to compare the cosine similarity.
It may cause the update speed of EATA to slow down slightly.
For PCL, we use the perturbation of normal distribution and dropout at the same time, where the normal distribution is the standard normal distribution.
For the QA task, the dropout rate we selected is 0.3; for the NER task, the dropout rate we selected is 0.1.
We tune the model with the learning rate in \{1e-5, 2e-5, 5e-5, 1e-4, 2e-4\}.
For the QA task, we select XLM-Roberta-base and XLM-Roberta-large \cite{conneau2020unsupervised} as our backbone; for the NER task, we choose BERT-base and BERT-large as our backbone.
We run all experiments three times with different random seeds and take the averaged value as the final experimental results.
We report the F1 and EM metrics for each QA dataset, F1 metric for NER datasets, and also provide the processing speed of the test-time methods in terms of samples per second, which is used to evaluate their efficiency.
All experiments are completed on NVIDIA RTX A5000 GPU.
Detailed data of all hyper-parameters are given in the Appendix~\ref{hyper}.


\subsection{Main Results on QA}
Table \ref{tab:qa} shows the results of all baselines and our method on the QA task.
We select XLM-Roberta-base and XLM-Roberta-large as the base model.
To train the source model on SQuAD with direct fine-tuning, we use the default training setup from \cite{hu2020xtreme}.
About robustness tuning, we select a strong baseline xTune \cite{zheng2021consistency} and train the base model under the default training setup.

\paragraph{Robustness Tuning}
xTune is an effective method to improve the robustness of the model.
With xTune, the generalization of the model will be significantly improved.
xTune is superior to the direct fine-tune model in all four datasets.
For XLM-Roberta-base, xTune is 5.8\% higher on EM 3.5\% higher on F1 score than direct fine-tuning on average; For XLM-Roberta-large, xTune is 4.8\% higher on EM 2.7\% higher on F1 score than direct fine-tuning on average.
\paragraph{Test Time Adaptation}
Tent and EATA perform well in speed but not in accuracy.
When using XLM-Roberta-base as the base model, the accuracy of Tent and EATA is even worse than that of direct forward.
This may be because of the insufficient discrimination ability of the model itself and the unstable output, resulting in the low quality of pseudo-labels, which makes the effect of TTA not ideal.
When using XLM-Roberta-large as the base model, both Tent and EATA have achieved good results.
The effect of TTA has been significantly improved after the quality of pseudo-labels improved.
OIL can achieve stable and considerable improvement based on the base model.
With exponentially smoothed teacher and stable output, the quality of OIL's pseudo-label is higher than that of Tent and EATA, which makes it possible that even when XLM-Roberta-base is used as the base model, OIL can obtain a significant performance improvement.
However, the inference speed of OIL is too slow.
In TTA, inference speed is a critical factor.
Compared with direct forward, the speed of OIL is about 20 times slower, which we cannot accept.
PCL can maintain high inference speed and high accuracy at the same time.
PCL accelerates the inference speed by nearly ten times while maintaining comparable accuracy as OIL.
PCL can obtain higher accuracy than Tent and EATA while keeping slightly faster than them. 
When XLM-Roberta-large is used as the base model, PCL can achieve an average of 5.1\% EM improvement and 2.4\% F1 improvement on four datasets.
PCL is about two times slower than direct forward, which is acceptable.
Test time adaptation can generally achieve considerable performance improvement based on the base model, but this will sacrifice certain inference speeds.
\paragraph{TTA with Robustness Tuning}
Robustness tuning and TTA are compatible, and the effect of TTA depends mainly on the model's performance.
If robustness tuning has a performance improvement based on the model itself, TTA can further improve the model's performance.
When using XLM-Roberta-large as the base model, PCL has increased EM and F1 by 5.2\% and 2.4\% based on xTune, 5.1\%, and 2.4\% based on direct fine-tuning.
Robustness tuning and TTA complement each other; when both are used at the same time, the accuracy of the model will be dramatically improved and get SOTA results.
\begin{table}[t]
\centering
\small
\caption{\label{tab:ner} Benchmarking results (\%) on NER for BERT-base and BERT-large. Context refers to the use of context based attack, entity refers to the use of entity based attack, and full refers to the simultaneous use of the two attack methods. We report the results of F1 scores in this table. \textbf{Bold}: the best results}
\vskip 0.15in
    \begin{tabular}{l c c c c c}
    \toprule
\textbf{Model} & \textbf{Context} & \textbf{Entity} & \textbf{Full} & \textbf{Avg.} & \textbf{Speed} \\
    \midrule
    bert-base & 82.38 & 59.59 & 54.68 & 65.55 & 1952.36 \\
    Tent & 82.33 & 61.16 & 57.72 & 67.07 & 579.68  \\
    EATA & 82.25 & 61.37 & 58.09 & 67.23 & 531.60 \\
    OIL & 82.07 & 61.53 & 56.96 & 66.85 & 236.94  \\
    PCL & \textbf{82.99} & \textbf{63.38} & \textbf{59.21} & \textbf{68.53} & 631.63\\
    \midrule
    bert-large & 83.11 & 62.44 & 57.78 & 67.77 & 1030.71 \\
    Tent & 82.60 & 63.68 & 59.03 & 68.43 & 273.03 \\
    EATA & 82.35 & 63.01 & 59.05 & 68.14 & 255.86\\
    OIL & 82.90 & 64.57 & 59.45 & 68.97 & 104.12 \\
    PCL & \textbf{83.28} & \textbf{65.63} & \textbf{62.09} & \textbf{70.33} & 294.44 \\
    \bottomrule
    \end{tabular}
\vskip -0.1in
\end{table}
\subsection{Main Results on NER}
Table \ref{tab:ner} shows the results of various baselines and our method on the RockNER dataset.
Tent and EATA can improve performance in NER tasks while maintaining good inference speed.
OIL can also improve the performance of the model. However, the improvement on this dataset is relatively limited, which is comparable to that of Tent and EATA, and the speed of OIL is still not fast enough, nearly ten times slower than that of direct forward.
PCL is slightly faster than Tent and EATA, about three times faster than OIL, and the effect is significantly better than theirs.
PCL can increase F1 value by 2.98\% based on BERT-base and 2.56\% based on BERT-large.
TTA has no pronounced effect on the context-level attachment on RockNER.
This may be because the context-level attack is not effective enough, resulting in the high accuracy of the model itself after the attack, and it is not easy to improve on this basis.
For entity-level attacks and full attacks, TTA has a very significant effect.
PCL achieves SOTA results on both inference speed and performance improvement.

\section{Analysis and Discussion}
\subsection{Online Learning}
\begin{figure}[t]
    \vskip 0.2in
    \centering
    \includegraphics[scale=0.35]{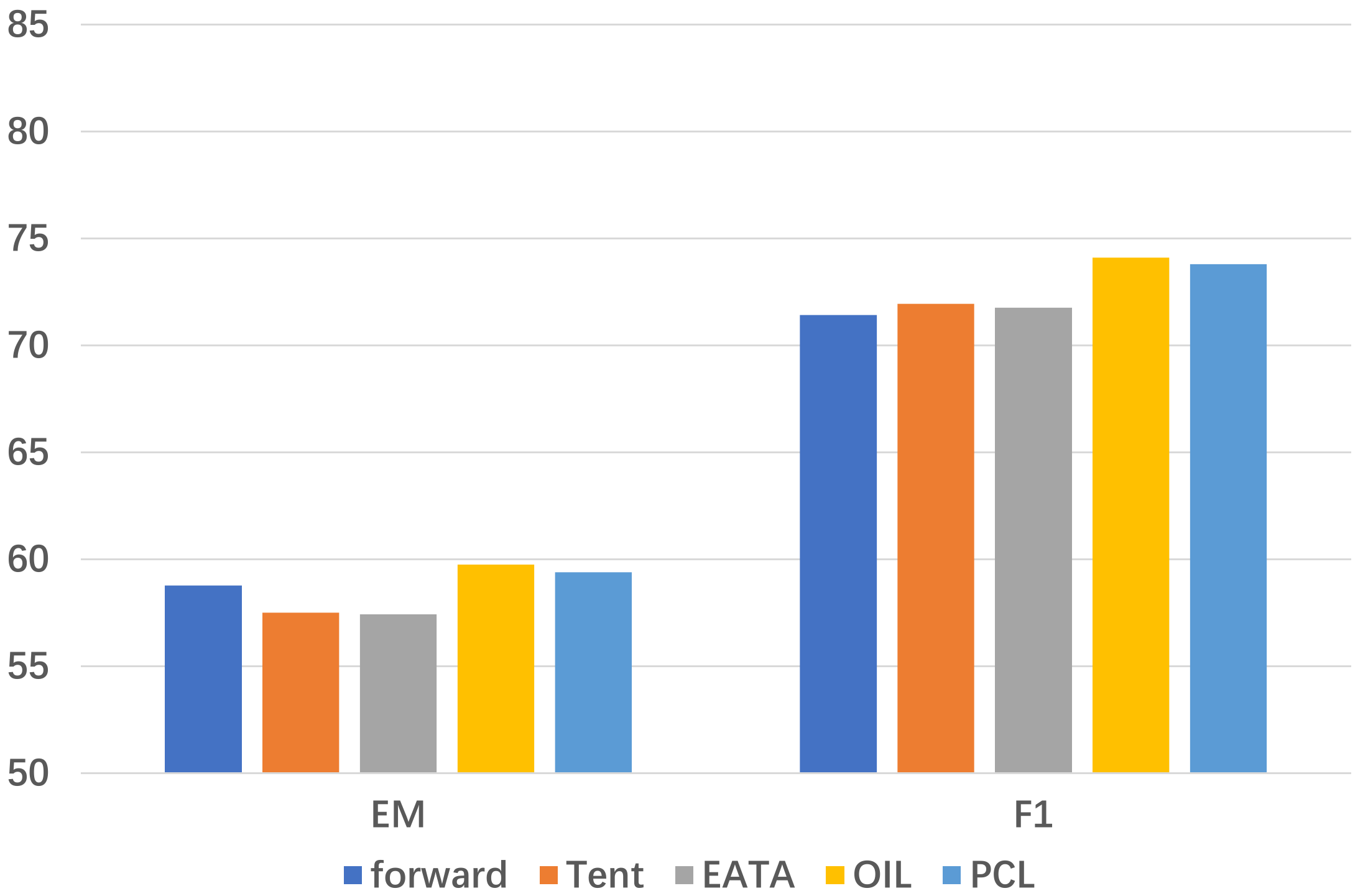}
    \caption{Results (\%) of different TTA methods and direct forward on XQuAD, where the base model is XLMR-base-xTune.}
    \label{fig:online-base}
    \vskip -0.2in
\end{figure}
\begin{figure}[t]
    \vskip 0.2in
    \centering
    \includegraphics[scale=0.35]{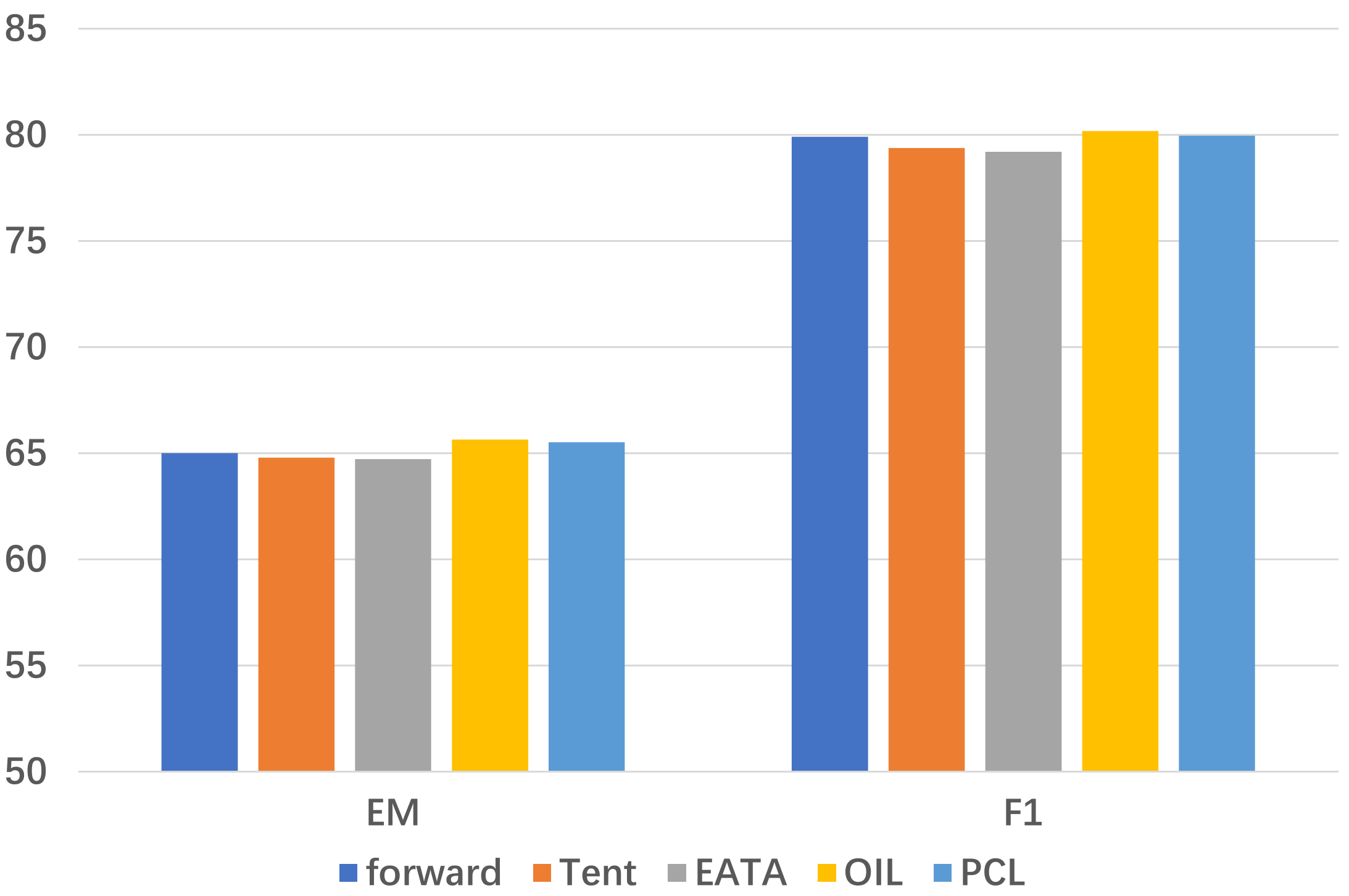}
    \caption{Results (\%) of different TTA methods and direct forward on XQuAD, where the base model is XLMR-large-xTune.}
    \label{fig:online-large}
    \vskip -0.2in
\end{figure}
Online learning is a typical setting in TTA. In this case, TTA will update the model without resetting it. 
Online is a more challenging and practical setting, requiring the model to adapt to changing distribution rather than a single target distribution.
Under this setting, TTA is more likely to output false pseudo-labels when the distribution changes, which makes the model accumulate errors during the test-time stage and is even prone to collapse.
We test the performance of each TTA method under the online setting on the XQuAD dataset.
Figure \ref{fig:online-base} and Figure \ref{fig:online-large} show the results of each TTA method when XLM-Roberta-base-xTune and XLM-Roberta-large-xTune are used as base models.
Under the online setting, the effects of both Tent and EATA are not ideal, and both have some performance degradation based on the base model.
OIL and PCL can still achieve performance improvement.
The performance improvement is minimal when XLM-Roberta-large-xTune is used as the base model.
When XLM-Roberta-base-xTune is used as the base model, the improvement of EM and F1 is more prominent.
The performance of OIL and PCL are equivalent, but the inference speed of OIL makes it not a good solution.

\subsection{Learning Rate}
\begin{figure}[t]
    \vskip 0.2in
    \centering
    \includegraphics[scale=0.35]{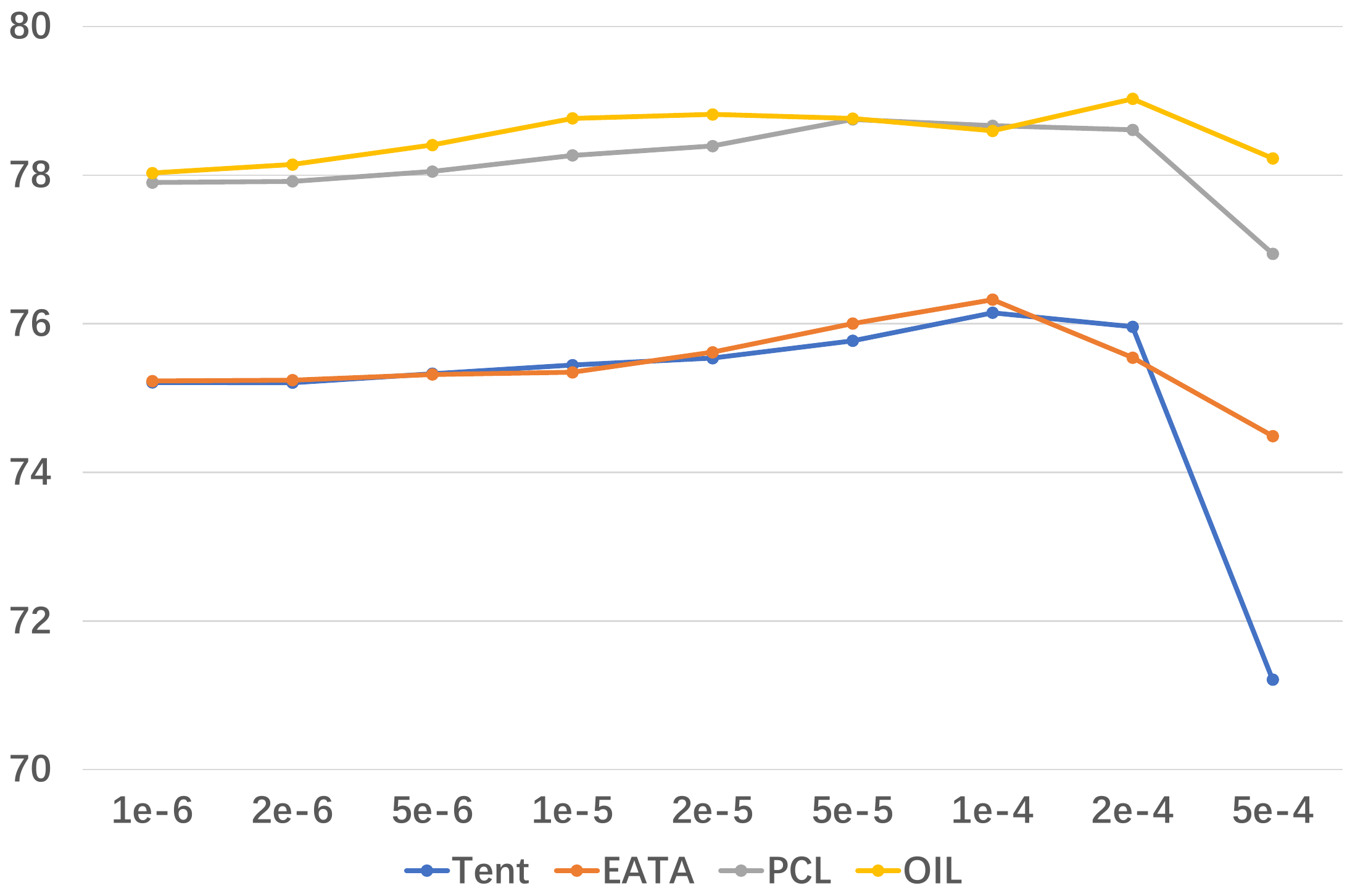}
    \caption{F1 scores of different TTA methods and different learning rates on NoiseQA-na, where the base model is XLM-Roberta-base.}
    \label{fig:lr}
    \vskip -0.2in
\end{figure}
The learning rate is a very important hyper-parameter in TTA, and different learning rates will significantly impact the effect of TTA.
When the learning rate is too low, the model parameters are not sufficiently updated, which will lead to the effect of TTA not being obvious.
When the learning rate is too big, the model easily collapses, and the model's performance is significantly reduced.
In actual application scenarios, it is unrealistic to select the learning rate in the test-time phase finely.
Thus, TTA's sensitivity to the learning rate is a significant factor in evaluating the effectiveness of a TTA method.
In order to verify the sensitivity of each baseline and PCL to the learning rate, we conduct experiments on the NoiseQA-na dataset with XLM-Roberta-base as the base model.
Figure \ref{fig:lr} shows the experimental results.
The overall trend of each method is the same. 
It can be seen from the figure that TTA has no apparent effect when the learning rate is relatively low.
With the increase in the learning rate, the model's performance improves.
After reaching the peak, continuing to increase the learning rate will make the performance of the model decline rapidly. 
Among them, Tent has the fastest decline speed.
Meanwhile, EATA, OIL, and PLC exhibit a relatively stable descent rate, making them more robust to learning rate than Tent.

\subsection{Perturbations}
\begin{table}[t]
\centering
\small
\caption{\label{tab:perturbation} F1 scores of different perturbing methods on RockNER.}
\vskip 0.15in
    \begin{tabular}{l c c c c}
    \toprule
\textbf{Model} & \textbf{Context} & \textbf{Entity} & \textbf{Full} & \textbf{Avg.} \\
    \midrule
    bert-base & 82.38 & 59.59 & 54.68 & 65.55  \\
    PCL & 82.99 & 63.38 & 59.21 & 68.53 \\
    PCL w/o norm & 82.84 & 62.17 & 58.70 & 67.90   \\
    PCL w/o dropout & 82.84 & 62.66 & 58.95 & 68.15  \\
    
    \midrule
    bert-large & 83.11 & 62.44 & 57.78 & 67.77  \\
    PCL & 83.28 & 65.63 & 62.09 & 70.33  \\
    PCL w/o norm  & 83.15 & 65.40 & 61.84 & 70.13  \\
    PCL w/o dropout & 83.24 & 65.01 & 61.98 & 70.08 \\
    
    \bottomrule
    \end{tabular}
\vskip -0.1in
\end{table}
The core of PCL is its perturbing method.
In order to study the effect of different perturbing methods, we use BERT-base and BERT-large as the base model and conduct experiments on the RockNER dataset.
Table \ref{tab:perturbation} shows the results under different perturbation methods.
The model's performance will significantly improve if the perturbation is added and back-propagated.
The method of adding normal distribution to the feature and the method of using dropout have good results, and the results are comparable.
When the two perturbation methods are used simultaneously, the model's performance will be further improved.
We believe that there are still better ways to add perturbations, such as using neural networks to generate the perturbations.
When the way to add perturbations improves, the performance of PCL will also improve.

\subsection{Dropout Rate}
\begin{figure}[t]
    \vskip 0.2in
    \centering
    \includegraphics[scale=0.35]{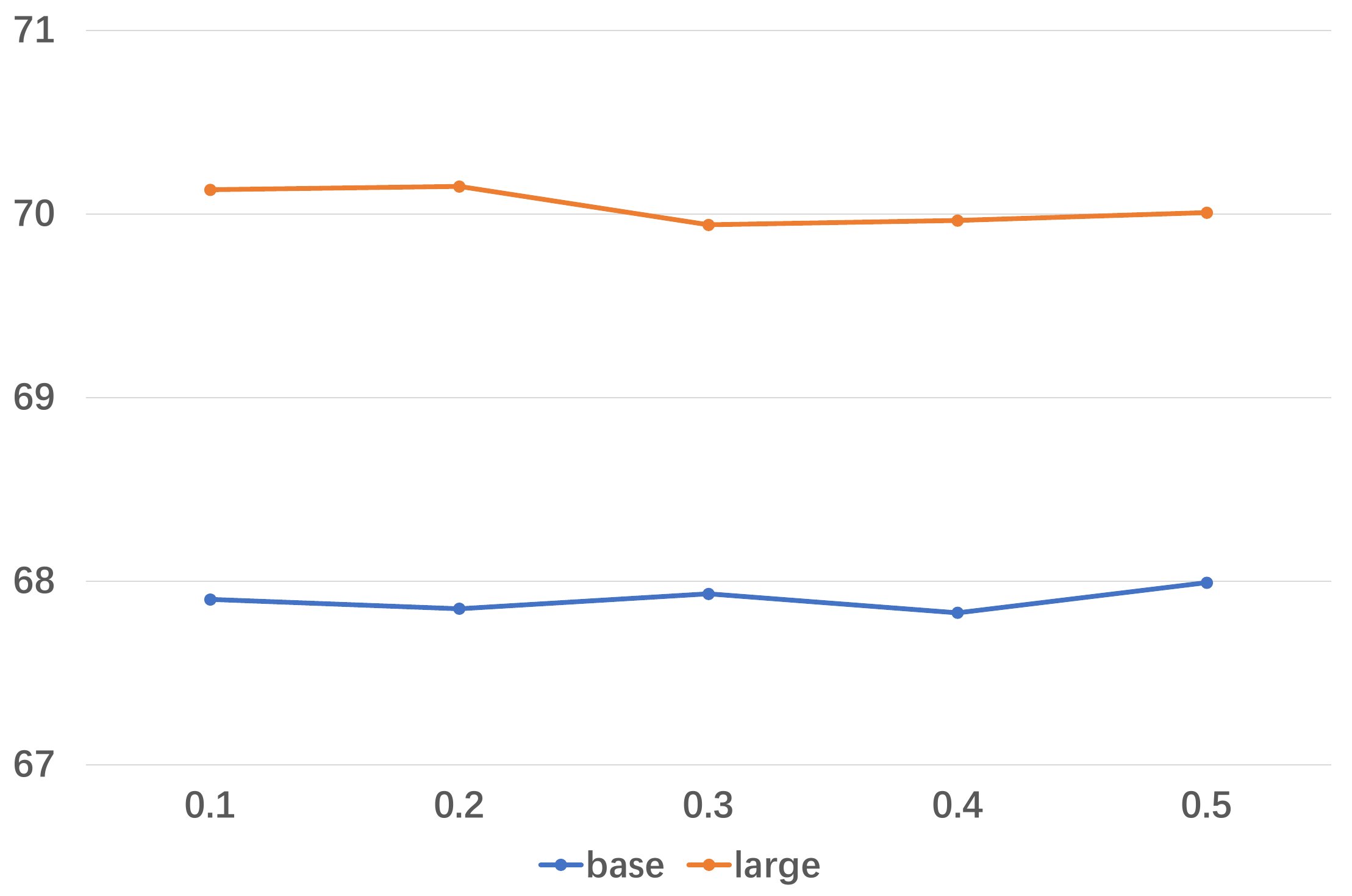}
    \caption{F1 scores of different dropout rates on RockNER.}
    \label{fig:drop}
    \vskip -0.2in
\end{figure}
When using dropout as the perturbing method, the dropout rate is also a very important hyper-parameter, directly determining how strong the perturbation is.
We hope PCL is not so sensitive to this hyper-parameter and maintains a good effect under different dropout rates.
In order to verify the robustness of PCL to the dropout rate, we use BERT-base and BERT-large as the base model and carry out experiments on the RockNER dataset.
The results are shown in figure \ref{fig:drop}. In general, PCL is robust to the dropout rate.
Under different dropout rates, PCL can perform well consistently. 
This may be because the feature layer itself is robust to dropout when the model is trained.
It can be seen from figure \ref{fig:drop} that when the dropout rate is slightly lower, the performance of PCL may be slightly better, which is also consistent with our core idea, i.e., maintaining the consistency of output under small perturbations.

\subsection{Error Analysis}
\begin{figure}[t]
\vskip 0.2in
    \centering
    \includegraphics[scale=0.35]{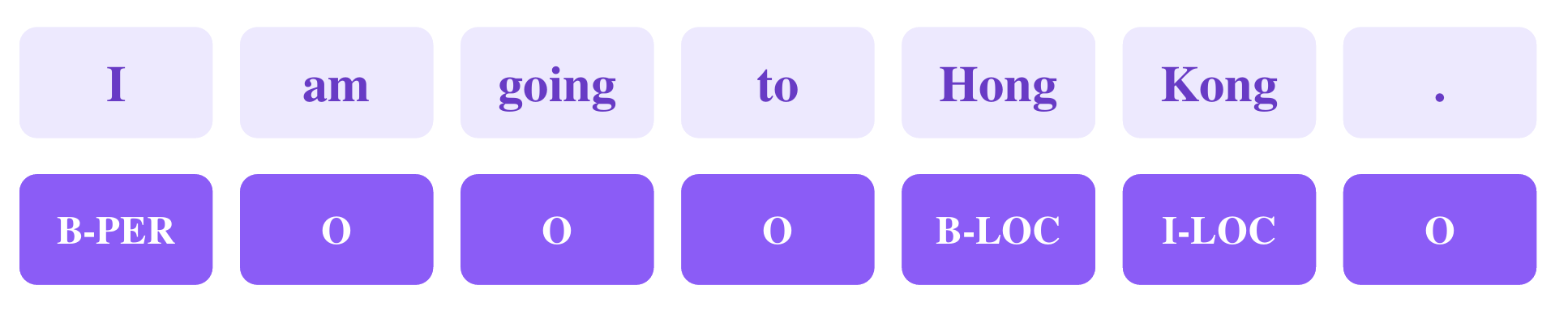}
    \caption{A brief example of NER entities.}
    \label{fig:error}
    \vskip -0.2in
\end{figure}
\begin{table}[t]
\caption{\label{tab:error} Results of error analysis. R $\xrightarrow{}$ W refers to right prediction for fine-tuning and wrong prediction for TTA, similarly as W $\xrightarrow{}$ R. Net Value refers to the difference between the two value.}
\vskip 0.15in
\centering
    \begin{tabular}{l c c c}
    \toprule
\textbf{Model} & \textbf{R $\xrightarrow{}$ W} &\textbf{W $\xrightarrow{}$ R} & \textbf{Net Value}\\
    \midrule
    Tent & 1038 & 1002 & -36  \\
    EATA & 1298 & 1363 & 65 \\
    OIL & 1007 & 1437 & 430  \\
    PCL & 985 & 1432 & 447 \\
    \bottomrule
    \end{tabular}
\vskip -0.1in
\end{table}
We conduct more fine-grained experiments to study the potential and drawbacks of PCL.
Specifically, we want to know how many samples that are mispredicted can be predicted correctly and how many samples that are predicted correctly will be mispredicted after using PCL compared with the direct fine-tuning model.
We extract all the entities in the experiment without considering the token labeled ``O".
If the model prediction is entirely correct, we believe this entity's prediction is correct.
As shown in Figure \ref{fig:error}, if the prediction result of ``I" is ``B-PER'', we believe that the entity's prediction is correct;
We conduct experiments on the RockNER dataset with BERT-base as the base model.
The results are shown in Table \ref{tab:error}.
The model's performance with Tent on entities will even be weakened, and EATA will slightly improve based on the base model.
The effect of OIL and PCL is significantly better than that of Tent and EATA.
We find that even though TTA can predict many entities with wrong predictions correct, it will also predict many entities with correct predictions wrong.
We believe that if the R $\xrightarrow{}$ W value can be reduced, the effect of TTA will be significantly increased.
We speculate that forgetting from TTA leads to incorrect prediction of samples that the model could have correctly predicted. Therefore, reducing forgetting in the test-time phase is a promising direction to further improve the performance of TTA.

\section{Conclusion}
In this paper, we analyze the problems of existing TTA methods in NLP tasks and propose a simple yet effective TTA method, PCL.
Our method turns off dropout in the trained model to avoid the performance degradation caused by dropout and then adds perturbation to the features and constrains the predictions' consistency with the original features to avoid overfitting caused by not using dropout.
Extensive experimental results demonstrate our method can achieve higher or comparable performance with less inference time than previous state-of-the-art TTA methods.
In future work, we will apply our PCL method to more challenging NLP tasks or settings, such as style-transfer text generation and cross-lingual syntax parsing.

\nocite{langley00}

\bibliography{example_paper}
\bibliographystyle{icml2023}

\newpage
\appendix
\onecolumn
\section{Hyper-parameters}
\label{hyper}
\begin{table*}[htb]
\centering
\caption{\label{tab:qa} Details of hyper-parameters.}
\vskip 0.15in
    \begin{tabular}{l c c c c c c c c c c c c}
    \toprule
\textbf{Models} &\textbf{NoiseQA} &\textbf{XQuAD} &\textbf{MLQA} &\textbf{RockNER} \\
    \midrule
     Tent & 
     \makecell[c]{LR(base): 2e-4 \\ LR(large): 2e-4 \\ BATCH\_SIZE: 8} & 
     \makecell[c]{LR(base): 2e-4 \\ LR(large): 2e-4 \\ BATCH\_SIZE: 8} & 
     \makecell[c]{LR(base): 1e-5 \\ LR(large): 1e-5 \\ BATCH\_SIZE: 8} & 
     \makecell[c]{LR(base): 5e-5 \\ LR(large): 1e-4 \\ BATCH\_SIZE: 16}\\

    \midrule
    EATA & 
    \makecell[c]{LR(base): 2e-4 \\ LR(large): 2e-4 \\ BATCH\_SIZE: 8\\$E_0: ln(1000)/2-1$ \\ $\beta:5e-4$ \\ $\epsilon: 0.05$}& 
    \makecell[c]{LR(base): 2e-4 \\ LR(large): 2e-4 \\ BATCH\_SIZE: 8\\$E_0: ln(1000)/2-1$ \\ $\beta:5e-4$ \\ $\epsilon: 0.05$}& 
    \makecell[c]{LR(base): 1e-5 \\ LR(large): 1e-5 \\ BATCH\_SIZE: 8\\$E_0: ln(1000)/2-1$ \\ $\beta:5e-4$ \\ $\epsilon: 0.05$}& 
    \makecell[c]{LR(base): 5e-5 \\ LR(large): 1e-4 \\ BATCH\_SIZE: 16\\$E_0: ln(1000)/2-1$ \\ $\beta:5e-4$ \\ $\epsilon: 0.05$}& \\
    \midrule
    OIL &
    \makecell[c]{LR(base): 2e-4 \\ LR(large): 2e-4 \\ BATCH\_SIZE: 8\\$K: 5$ \\ $\gamma:0.5$ \\ $\alpha: 0.99$ \\ $\beta: 1$}& 
    \makecell[c]{LR(base): 2e-4 \\ LR(large): 2e-4 \\ BATCH\_SIZE: 8\\$K: 5$ \\ $\gamma:0.5$ \\ $\alpha: 0.99$ \\ $\beta: 1$}& 
    \makecell[c]{LR(base): 1e-5 \\ LR(large): 1e-5 \\ BATCH\_SIZE: 8\\$K: 3$ \\ $\gamma:\infty$ \\ $\alpha: 0.99$ \\ $\beta: 1$}& 
    \makecell[c]{LR(base): 5e-5 \\ LR(large): 1e-4 \\ BATCH\_SIZE: 16\\$K: 16$ \\ $\gamma:0.5$ \\ $\alpha: 0.99$ \\ $\beta: 1$}& \\
    \midrule
    PCL &
    \makecell[c]{LR(base): 2e-4 \\ LR(large): 2e-4 \\ BATCH\_SIZE: 8} & 
    \makecell[c]{LR(base): 2e-4 \\ LR(large): 2e-4 \\ BATCH\_SIZE: 8} & 
    \makecell[c]{LR(base): 1e-5 \\ LR(large): 1e-5 \\ BATCH\_SIZE: 8} & 
    \makecell[c]{LR(base): 5e-5 \\ LR(large): 1e-4 \\ BATCH\_SIZE: 16} & \\
    \bottomrule
    \end{tabular}
\vskip -0.1in
\end{table*}
\section{Detailed Results}
\begin{table*}[htb]
\centering
\caption{Detailed results (\%) on NoiseQA-syn.}
\vskip 0.15in
    \begin{tabular}{l c c c c c c c c c c c }
    \toprule
\multirow{2}{*}{\textbf{Models}} & \multicolumn{2}{c}{\textbf{asr}} & \multicolumn{2}{c}{\textbf{keyboard}} & \multicolumn{2}{c}{\textbf{translation}}& \multicolumn{2}{c}{\textbf{avg}}\\
\cmidrule(r){2-3} \cmidrule(r){4-5} \cmidrule(r){6-7} \cmidrule(r){8-9}
 &\textbf{EM} & \textbf{F1} & \textbf{EM} &\textbf{F1} & \textbf{EM} & \textbf{F1} & \textbf{EM} & \textbf{F1}\\
    \midrule
    xlmr-base & 67.14 & 79.95 & 64.71 & 76.29 & 67.98 & 79.66 & 66.64 & 78.67\\
    xTune & 71.51 & 82.62 & 70.08 & 80.02 & 71.18 & 81.83 & 70.95 & 81.52\\
    \midrule
    Tent & 67.20 & 78.90 & 63.47 & 74.51 & 66.16 & 77.78 & 65.61 & 77.06\\
    EATA & 66.92 & 78.67 & 62.86 & 73.95 & 65.91 & 77.58 & 65.23 & 76.73\\
    OIL & 69.83 & 81.28 & 66.81 & 77.50 & 69.02 & 80.18 & 68.55 & 79.65\\
    PCL & 70.06 & 81.53 & 66.39 & 77.24 & 68.88 & 80.07 & 68.44 & 79.61\\
    \midrule
    xTune+Tent & 71.18 & 81.95 & 69.58 & 79.75 & 70.00 & 80.70 & 70.25 & 80.80\\
    xTune+EATA & 70.78 & 81.58 & 69.22 & 79.36 & 69.50 & 80.33 & 69.83 & 80.42\\
    xTune+OIL & 72.35 & 82.85 & 71.32 & 81.03 & 70.86 & 81.58 & 71.51 & 81.82\\
    xTune+PCL & 72.16 & 82.77 & 71.51 & 81.09 & 70.90 & 81.77 & 71.52 & 81.88\\
    \midrule
    \midrule
    xlmr-large & 53.45 & 72.47 & 72.94 & 84.66 & 70.25 & 82.60 & 65.55 & 79.91 \\
    xTune & 58.57 & 75.83 & 76.47 & 87.36 & 75.29 & 86.32 & 70.11 & 83.17\\
    \midrule
    Tent & 71.34 & 83.37 & 72.02 & 83.49 & 69.89 & 82.07 & 71.08 & 82.97\\
    EATA & 71.88 & 83.69 & 72.30 & 83.49 & 69.41 & 81.62 & 71.20 & 82.93\\
    OIL & 69.83 & 83.43 & 73.22 & 85.01 & 70.06 & 82.61 & 71.03 & 83.68\\
    PCL & 71.48 & 83.99 & 73.33 & 84.57 & 70.76 & 82.69 & 71.86 & 83.75\\
    \midrule
    xTune+Tent & 75.77 & 86.52 & 76.41 & 86.84 & 74.99 & 85.60 & 75.72 & 86.32\\
    xTune+EATA & 75.63 & 86.40 & 76.39 & 86.68 & 74.68 & 85.29 & 75.56 & 86.12\\
    xTune+OIL & 75.69 & 86.79 & 76.81 & 87.25 & 75.32 & 86.13 & 75.94 & 86.72\\
    xTune+PCL & 75.85 & 86.69 & 76.83 & 87.22 & 75.63 & 86.18 & 76.11 & 86.69\\
    \bottomrule
    \end{tabular}
\vskip -0.1in
\end{table*}

\begin{table*}[htb]
\centering
\caption{Detailed results (\%) on NoiseQA-na.}
\vskip 0.15in
    \begin{tabular}{l c c c c c c c c c c c }
    \toprule
\multirow{2}{*}{\textbf{Models}} & \multicolumn{2}{c}{\textbf{asr}} & \multicolumn{2}{c}{\textbf{keyboard}} & \multicolumn{2}{c}{\textbf{translation}}& \multicolumn{2}{c}{\textbf{avg}}\\
\cmidrule(r){2-3} \cmidrule(r){4-5} \cmidrule(r){6-7} \cmidrule(r){8-9}
 &\textbf{EM} & \textbf{F1} & \textbf{EM} &\textbf{F1} & \textbf{EM} & \textbf{F1} & \textbf{EM} & \textbf{F1}\\
    \midrule
    xlmr-base & 57.98 & 70.79 & 71.01 & 82.41 & 69.24 & 80.56 & 66.25 & 77.91\\
    xTune & 63.78 & 75.16 & 73.95 & 84.59 & 71.43 & 82.20 & 69.75 & 80.66\\
    \midrule
    Tent & 59.33 & 70.65 & 68.96 & 80.37 & 66.89 & 78.12 & 65.06 & 76.38\\
    EATA & 58.68 & 69.89 & 68.68 & 80.33 & 66.58 & 77.92 & 64.65 & 76.05\\
    OIL & 63.56 & 74.43 & 70.81 & 82.28 & 68.88 & 80.11 & 67.75 & 78.94\\
    PCL & 62.27 & 73.52 & 71.43 & 82.74 & 68.52 & 79.82 & 67.40 & 78.69\\
    \midrule
    xTune+Tent & 65.35 & 74.75 & 73.53 & 83.87 & 70.67 & 81.51 & 69.85 & 80.04\\
    xTune+EATA & 65.04 & 74.59 & 73.08 & 83.64 & 70.31 & 81.19 & 69.48 & 79.81\\
    xTune+OIL & 65.85 & 75.98 & 73.92 & 84.35 & 71.29 & 81.96 & 70.35 & 80.76\\
    xTune+PCL & 66.27 & 76.40 & 74.23 & 84.82 & 71.54 & 82.24 & 70.68 & 81.15\\
    \midrule
    \midrule
    xlmr-large & 46.13 & 64.62 & 73.53 & 85.86 & 72.86 & 84.63 & 64.17 & 78.37 \\
    xTune & 49.41 & 66.78 & 76.47 & 88.12 & 75.71 & 86.72 & 67.20 & 80.54\\
    \midrule
    Tent & 61.18 & 73.76 & 73.75 & 85.40 & 70.78 & 82.80 & 68.57 & 80.65\\
    EATA & 61.71 & 73.92 & 73.47 & 85.15 & 71.12 & 82.76 & 68.77 & 80.61\\
    OIL & 61.18 & 74.78 & 74.65 & 86.26 & 72.94 & 84.52 & 69.59 & 81.85\\
    PCL & 61.88 & 74.96 & 73.98 & 86.22 & 73.33 & 84.53 & 69.73 & 81.90\\
    \midrule
    xTune+Tent & 66.81 & 77.64 & 77.87 & 88.05 & 75.43 & 86.10 & 73.37 & 83.93\\
    xTune+EATA & 66.89 & 77.92 & 77.31 & 87.73 & 75.41 & 85.98 & 73.20 & 83.88\\
    xTune+OIL & 66.81 & 78.32 & 77.54 & 88.31 & 76.22 & 86.81 & 73.52 & 84.48\\
    xTune+PCL & 67.70 & 78.38 & 77.82 & 88.51 & 76.16 & 86.85 & 73.89 & 84.58\\
    \bottomrule
    \end{tabular}
\vskip -0.1in
\end{table*}

\begin{table*}[htb]
\centering
\caption{Detailed results (\%) on XQuAD (EM).}
\vskip 0.15in
    \begin{tabular}{l c c c c c c c c c c c c}
    \toprule
\textbf{Models} & en & es & de & el & ru & tr & ar & vi & th & zh & hi & \textbf{avg} \\
    \midrule
    xlmr-base & 72.52 & 58.07 & 58.15 & 54.54 & 57.56 & 50.92 & 48.99 & 55.38 & 55.13 & 51.60 & 48.66 & 55.59\\
    xTune & 74.71 & 60.67 & 62.52 & 58.15 & 61.18 & 54.29 & 52.02 & 55.21 & 59.83 & 53.87 & 54.45 & 58.81\\
    \midrule
   Tent & 70.00 & 56.50 & 54.54 & 53.45 & 54.57 & 47.42 & 45.69 & 52.16 & 54.45 & 54.54 & 48.24 & 53.78\\
    EATA & 69.69 & 56.22 & 54.79 & 53.47 & 54.34 & 48.24 & 44.59 & 51.68 & 57.14 & 55.18 & 48.10 & 53.95\\
    OIL & 72.18 & 60.00 & 58.26 & 56.19 & 57.48 & 51.99 & 48.88 & 55.18 & 62.61 & 59.72 & 51.43 & 57.63\\
    PCL & 72.68 & 59.18 & 58.03 & 56.10 & 58.09 & 52.37 & 49.43 & 55.29 & 57.82 & 55.73 & 51.27 & 56.91\\
    \midrule
    xTune+Tent & 73.89 & 60.11 & 60.06 & 57.06 & 58.66 & 52.89 & 49.61 & 54.65 & 59.80 & 55.21 & 53.19 & 57.74\\
    xTune+EATA & 73.64 & 59.36 & 59.41 & 56.61 & 58.26 & 52.66 & 49.44 & 54.34 & 60.00 & 56.08 & 53.11 & 57.54\\
    xTune+OIL & 73.89 & 61.23 & 62.55 & 59.50 & 60.17 & 55.71 & 52.16 & 57.28 & 63.70 & 57.28 & 54.96 & 59.86\\
    xTune+PCL & 74.87 & 61.23 & 62.24 & 59.05 & 60.53 & 55.49 & 52.30 & 56.89 & 62.32 & 56.78 & 54.73 & 59.68\\
    \midrule
    \midrule
    xlmr-large & 75.38 & 63.87 & 62.52 & 63.78 & 63.53 & 59.92 & 58.57 & 61.60 & 64.62 & 61.51 & 59.33 & 63.15\\
    xTune & 77.82 & 65.80 & 67.31 & 66.47 & 65.71 & 61.68 & 60.50 & 63.28 & 63.78 & 59.41 & 63.28 & 65.00\\
    \midrule
    Tent & 74.17 & 62.66 & 62.13 & 61.04 & 61.65 & 57.90 & 57.28 & 59.02 & 65.46 & 65.57 & 58.10 & 62.27\\
    EATA & 73.98 & 62.77 & 62.18 & 61.12 & 61.76 & 58.10 & 56.72 & 58.85 & 65.94 & 66.13 & 57.96 & 62.32\\
    OIL & 75.91 & 63.78 & 62.72 & 62.61 & 63.64 & 59.86 & 58.66 & 61.26 & 68.99 & 68.74 & 60.53 & 64.24\\
    PCL & 75.63 & 64.03 & 62.63 & 63.84 & 63.64 & 59.86 & 58.71 & 60.92 & 66.64 & 65.69 & 60.22 & 63.80\\
    \midrule
    xTune+Tent & 78.04 & 66.41 & 66.69 & 65.15 & 64.59 & 61.74 & 60.20 & 62.94 & 67.11 & 61.85 & 62.77 & 65.23\\
    xTune+EATA & 77.62 & 66.08 & 66.50 & 64.51 & 64.31 & 61.29 & 59.94 & 62.41 & 67.31 & 61.85 & 62.61 & 64.95\\
    xTune+OIL & 78.32 & 67.14 & 68.21 & 65.85 & 65.32 & 62.13 & 60.64 & 63.89 & 68.21 & 63.31 & 63.22 & 66.02\\
    xTune+PCL & 78.35 & 67.00 & 66.92 & 66.30 & 65.10 & 62.18 & 60.76 & 64.15 & 67.31 & 62.46 & 63.42 & 65.81\\
    \bottomrule
    \end{tabular}
\vskip -0.1in
\end{table*}

\begin{table*}[htb]
\centering
\caption{Detailed results (\%) on XQuAD (F1).}
\vskip 0.15in
    \begin{tabular}{l c c c c c c c c c c c c}
    \toprule
\textbf{Models} & en & es & de & el & ru & tr & ar & vi & th & zh & hi & \textbf{avg} \\
    \midrule
    xlmr-base & 83.73 & 76.71 & 74.56 & 72.56 & 74.30 & 67.88 & 66.06 & 73.80 & 66.38 & 63.52 & 66.23 & 71.43\\
    xTune & 85.14 & 77.76 & 77.80 & 75.69 & 76.62 & 70.54 & 69.05 & 74.92 & 68.91 & 64.14 & 70.90 & 73.77\\
    \midrule
    Tent & 81.49 & 74.37 & 70.31 & 69.70 & 71.30 & 63.56 & 62.86 & 71.76 & 64.82 & 64.95 & 63.42 & 68.96\\
    EATA & 81.23 & 73.93 & 70.22 & 69.62 & 71.02 & 63.72 & 62.02 & 71.24 & 67.06 & 64.38 & 62.91 & 68.85\\
    OIL & 83.41 & 77.31 & 73.91 & 72.42 & 74.04 & 67.76 & 65.92 & 73.81 & 71.58 & 68.61 & 67.30 & 72.37\\
    PCL & 83.78 & 76.77 & 74.41 & 72.45 & 74.48 & 68.20 & 66.27 & 73.79 & 68.22 & 65.91 & 66.98 & 71.93\\
    \midrule
    xTune+Tent & 84.30 & 76.72 & 74.68 & 73.40 & 74.53 & 68.22 & 66.74 & 74.28 & 68.36 & 64.12 & 68.52 & 72.17\\
    xTune+EATA & 84.10 & 75.89 & 74.31 & 72.85 & 74.25 & 67.87 & 66.46 & 74.02 & 68.28 & 64.49 & 68.39 & 71.90\\
    xTune+OIL & 84.72 & 77.79 & 77.33 & 75.40 & 76.06 & 71.49 & 69.44 & 76.43 & 72.02 & 66.24 & 71.21 & 74.38\\
    xTune+PCL & 85.21 & 77.39 & 77.08 & 75.54 & 76.10 & 70.41 & 69.63 & 75.91 & 70.76 & 65.65 & 70.99 & 74.06\\
    \midrule
    \midrule
    xlmr-large & 86.96 & 82.57 & 80.20 & 81.27 & 79.75 & 76.58 & 75.71 & 80.73 & 75.26 & 71.02 & 76.45 & 78.77\\
    xTune & 88.96 & 83.56 & 83.26 & 82.64 & 82.08 & 77.29 & 77.51 & 82.47 & 73.62 & 68.67 & 78.92 & 79.91\\
    \midrule
    Tent & 85.95 & 81.00 & 79.27 & 79.09 & 78.61 & 74.75 & 74.23 & 78.76 & 75.41 & 73.70 & 74.66 & 77.77\\
    EATA & 85.79 & 80.73 & 79.12 & 78.71 & 78.67 & 74.51 & 73.66 & 78.52 & 75.55 & 73.96 & 74.22 & 77.59\\
    OIL & 87.03 & 82.43 & 79.81 & 80.49 & 80.43 & 76.13 & 75.70 & 80.41 & 77.89 & 76.24 & 76.89 & 79.40\\
    PCL & 86.97 & 82.15 & 79.90 & 80.76 & 80.01 & 76.09 & 75.69 & 80.14 & 76.60 & 73.94 & 76.45 & 78.97\\
    \midrule
    xTune+Tent & 88.46 & 82.90 & 82.44 & 81.81 & 81.04 & 77.02 & 76.65 & 81.71 & 75.81 & 69.51 & 78.05 & 79.58\\
    xTune+EATA & 88.27 & 82.47 & 82.13 & 81.17 & 80.77 & 76.72 & 76.43 & 81.43 & 75.58 & 69.22 & 77.85 & 79.28\\
    xTune+OIL & 89.05 & 83.72 & 83.27 & 82.32 & 81.62 & 77.45 & 77.34 & 82.61 & 76.61 & 70.69 & 79.07 & 80.34\\
    xTune+PCL & 88.79 & 83.40 & 82.85 & 82.39 & 81.50 & 77.46 & 77.44 & 82.54 & 75.62 & 70.04 & 78.80 & 80.08\\
    \bottomrule
    \end{tabular}
\vskip -0.1in
\end{table*}

\begin{table*}[htb]
\centering
\caption{Detailed results (\%) on MLQA (EM).}
\vskip 0.15in
    \begin{tabular}{l c c c c c c c c c c c c}
    \toprule
    \textbf{Models} & en & es & de & ar & hi & vi & zh & \textbf{avg} \\
    \midrule
    xlmr-base & 67.58 & 50.20 & 47.93 & 38.20 & 42.90 & 46.28 & 36.77 & 47.12\\
    xTune & 69.13 & 51.84 & 50.52 & 39.23 & 47.25 & 48.86 & 42.24 & 49.87\\
    \midrule
    Tent & 65.06 & 47.24 & 45.49 & 34.12 & 39.81 & 43.72 & 33.85 & 44.18\\
    EATA & 64.84 & 47.26 & 45.30 & 33.71 & 39.47 & 43.31 & 33.99 & 43.98\\
    OIL & 68.08 & 50.23 & 48.22 & 38.51 & 44.29 & 47.15 & 38.81 & 47.90\\
    PCL & 67.72 & 50.16 & 48.16 & 38.16 & 43.62 & 46.44 & 37.38 & 47.38\\
    \midrule
    xTune+Tent & 68.61 & 50.54 & 48.65 & 37.49 & 44.52 & 47.19 & 40.39 & 48.20\\
    xTune+EATA & 68.37 & 50.50 & 48.69 & 36.91 & 43.89 & 46.99 & 40.45 & 47.97\\
    xTune+OIL & 69.70 & 52.08 & 50.47 & 40.60 & 48.55 & 49.07 & 42.81 & 50.47\\
    xTune+PCL & 69.44 & 52.12 & 50.68 & 39.51 & 47.55 & 48.90 & 42.45 & 50.09\\
    \midrule
    \midrule
    xlmr-large & 70.14 & 56.33 & 55.88 & 46.56 & 52.75 & 52.59 & 42.87 & 53.87\\
    xTune & 72.89 & 58.04 & 56.87 & 48.12 & 55.41 & 54.47 & 48.32 & 56.30\\
    \midrule
    Tent & 69.57 & 54.74 & 53.99 & 43.77 & 50.47 & 51.36 & 41.42 & 52.19\\
    EATA & 69.49 & 54.69 & 53.91 & 43.76 & 50.37 & 51.20 & 41.59 & 52.15\\
    OIL & 70.71 & 56.42 & 55.91 & 46.57 & 52.56 & 52.59 & 44.59 & 54.19\\
    PCL & 70.53 & 56.41 & 55.79 & 46.68 & 52.56 & 52.58 & 43.48 & 54.00\\
    \midrule
    xTune+Tent & 72.68 & 57.02 & 56.42 & 46.83 & 54.03 & 53.59 & 47.19 & 55.39\\
    xTune+EATA & 72.62 & 56.98 & 56.50 & 46.70 & 53.96 & 53.56 & 47.17 & 55.36\\
    xTune+OIL & 73.05 & 57.74 & 57.07 & 47.98 & 55.22 & 54.32 & 48.08 & 56.21\\
    xTune+PCL & 73.12 & 57.84 & 56.93 & 48.04 & 55.52 & 54.34 & 48.26 & 56.29\\
    \bottomrule
    \end{tabular}
\vskip -0.1in
\end{table*}

\begin{table*}[htb]
\centering
\caption{Detailed results (\%) on MLQA (F1).}
\vskip 0.15in
    \begin{tabular}{l c c c c c c c c c c c c}
    \toprule
\textbf{Models} & en & es & de & ar & hi & vi & zh & \textbf{avg} \\
    \midrule
    xlmr-base & 80.45 & 67.68 & 62.53 & 57.12 & 61.02 & 67.06 & 61.01 & 65.27\\
    xTune & 81.75 & 69.56 & 65.41 & 59.03 & 64.37 & 69.62 & 64.54 & 67.76\\
    \midrule
    Tent & 78.15 & 64.97 & 60.29 & 52.97 & 57.46 & 64.28 & 57.98 & 62.30\\
    EATA & 77.90 & 64.87 & 60.07 & 52.53 & 56.94 & 63.85 & 58.01 & 62.02\\
    OIL & 80.62 & 67.54 & 62.46 & 57.23 & 61.43 & 67.48 & 62.03 & 65.54\\
    PCL & 80.40 & 67.36 & 62.49 & 56.98 & 61.07 & 66.95 & 61.33 & 65.23\\
    \midrule
    xTune+Tent & 80.93 & 68.05 & 63.54 & 56.61 & 61.21 & 67.70 & 62.77 & 65.83\\
    xTune+EATA & 80.70 & 67.96 & 63.58 & 55.95 & 60.44 & 67.41 & 62.83 & 65.55\\
    xTune+OIL & 82.04 & 69.58 & 65.25 & 60.14 & 65.37 & 69.89 & 65.18 & 68.21\\
    xTune+PCL & 81.73 & 69.69 & 65.43 & 59.29 & 64.22 & 69.58 & 64.67 & 67.80\\
    \midrule
    \midrule
    xlmr-large & 83.46 & 74.63 & 70.68 & 66.74 & 70.85 & 74.00 & 67.67 & 72.58\\
    xTune & 85.40 & 75.57 & 71.82 & 67.89 & 73.05 & 75.79 & 70.78 & 74.33\\
    \midrule
    Tent & 82.74 & 73.04 & 69.16 & 64.19 & 68.34 & 72.85 & 66.27 & 70.94\\
    EATA & 82.66 & 72.96 & 69.02 & 64.06 & 68.20 & 72.74 & 66.32 & 70.85\\
    OIL & 83.76 & 74.60 & 70.56 & 66.73 & 70.91 & 74.01 & 68.48 & 72.72\\
    PCL & 83.60 & 74.58 & 70.47 & 66.73 & 70.71 & 73.92 & 67.95 & 72.56\\
    \midrule
    xTune+Tent & 85.00 & 74.79 & 71.17 & 66.84 & 71.79 & 74.97 & 69.97 & 73.50\\
    xTune+EATA & 84.90 & 74.70 & 71.17 & 66.61 & 71.71 & 74.88 & 69.95 & 73.42\\
    xTune+OIL & 85.45 & 75.41 & 71.89 & 67.67 & 73.00 & 75.63 & 70.59 & 74.23\\
    xTune+PCL & 85.44 & 75.47 & 71.81 & 67.74 & 73.08 & 75.63 & 70.72 & 74.27\\
    \bottomrule
    \end{tabular}
\vskip -0.1in
\end{table*}


\end{document}